\documentclass[10pt,twocolumn,letterpaper]{article}

\usepackage{cvpr}
\usepackage{times}
\usepackage{epsfig}
\usepackage{graphicx}
\usepackage{amsmath}
\usepackage{amssymb}
\usepackage{caption}
\usepackage{subcaption}
\usepackage{fix2col}
\usepackage{xcolor}
\usepackage{diagbox}
\usepackage{xspace}
\usepackage{tabularx,booktabs}
\usepackage[nocompress]{cite}

\newcommand{\reffig}[1]{Fig.~\ref{fig:#1}}
\newcommand{\refsec}[1]{Sec.~\ref{sec:#1}}

\newcommand{\reftbl}[1]{Table~\ref{tbl:#1}}

\newcommand{\lblfig}[1]{\label{fig:#1}}
\newcommand{\lblsec}[1]{\label{sec:#1}}

\newcommand{\lbltbl}[1]{\label{tbl:#1}}

\newcommand{\ignorethis}[1]{}

\newcommand{\LOneLoss}{$L_1$ }
\newcommand{\ck}{Chen and Koltun~\cite{chen2017photographic}\xspace}
\newcommand{\pp}{pix2pix\xspace}
\newcommand{\s}{\mathbf{s}}
\newcommand{\x}{\mathbf{x}}
\newcommand{\ours}{ours\xspace}
\newcolumntype{Y}{>{\centering\arraybackslash}X}

\usepackage[pagebackref=true,breaklinks=true,letterpaper=true,colorlinks,bookmarks=false]{hyperref}

\cvprfinalcopy
\def\cvprPaperID{0000}

\ifcvprfinal\fi
\begin{document}

\title{High-Resolution Image Synthesis and Semantic Manipulation with Conditional GANs}

\author{Ting-Chun Wang\textsuperscript{1} \quad Ming-Yu Liu\textsuperscript{1} \quad Jun-Yan Zhu\textsuperscript{2} \quad Andrew Tao\textsuperscript{1} \quad Jan Kautz\textsuperscript{1} \quad Bryan Catanzaro\textsuperscript{1}
\\
\textsuperscript{1}NVIDIA Corporation \qquad \textsuperscript{2}UC Berkeley}

\twocolumn[{%
\renewcommand\twocolumn[1][]{#1}%
\maketitle
  \centering
  \vspace{-.2in}
  \includegraphics[width=\linewidth]{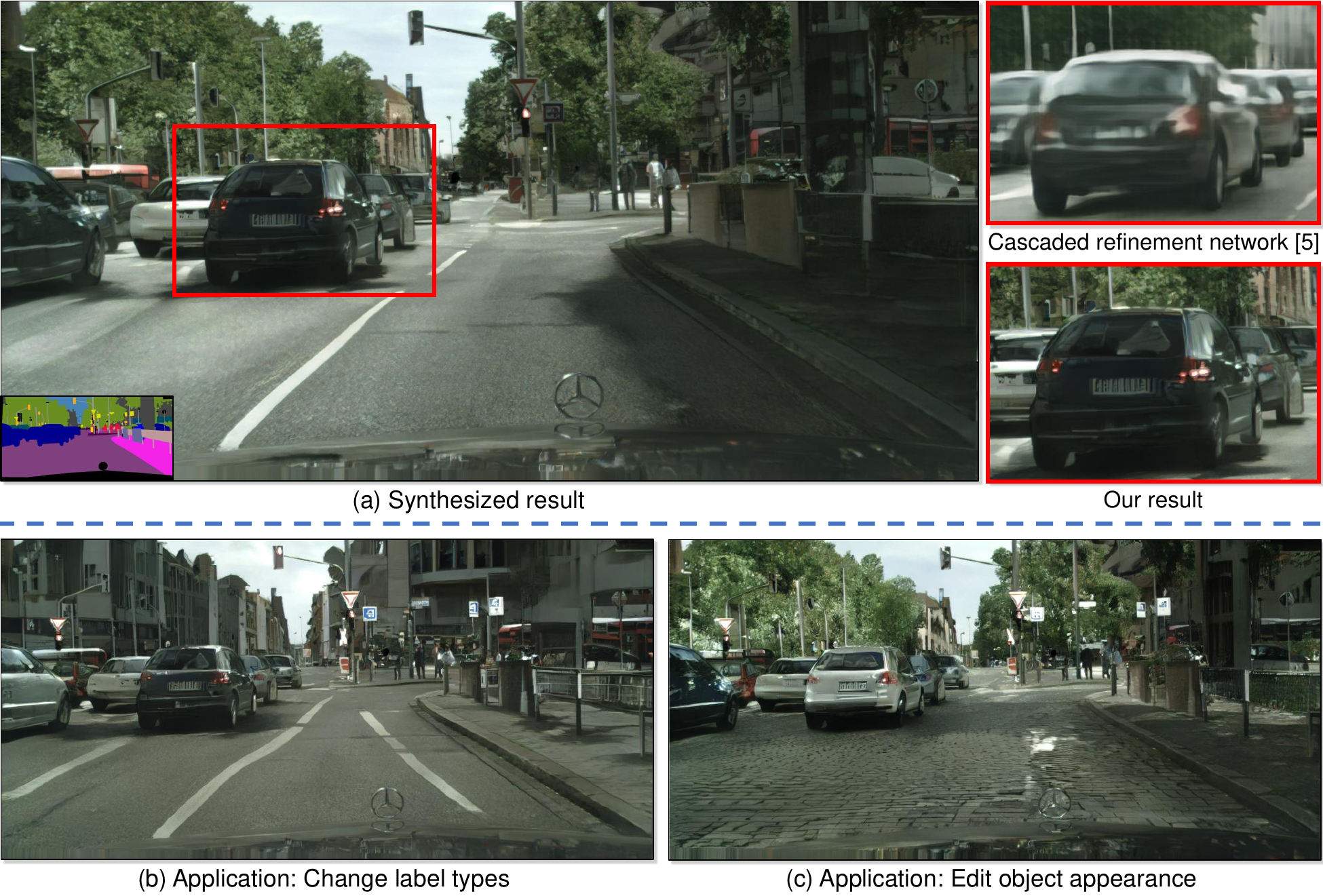}
  \vspace{-.25in}
  \captionof{figure}{We propose a generative adversarial framework for synthesizing $2048\times 1024$ images from semantic label maps (lower left corner in (a)). Compared to previous work~\cite{chen2017photographic}, our results express more natural textures and details. (b) We can change labels in the original label map to create new scenes, like replacing trees with buildings. (c) Our framework also allows a user to edit the appearance of individual objects in the scene, e.g.\ changing the color of a car or the texture of a road.  Please visit our \href{https://tcwang0509.github.io/pix2pixHD/}{website} for more side-by-side comparisons as well as interactive editing demos.}
  \vspace{.15in}
  \lblfig{teaser}
}]

\maketitle

\newif\ifsubmit
\submitfalse
\ifsubmit

\newcommand{\jy}[1]{}
\newcommand{\my}[1]{}
\newcommand{\JK}[1]{}
\else
\newcommand{\JK}[1]{\textcolor{red}{Jan: #1}}
\newcommand{\revJK}[1]{\textcolor{red}{#1}}
\newcommand{\jy}[1]{\textcolor{blue}{J.-Y.Z: #1}}
\newcommand{\revjy}[1]{\textcolor{blue}{#1}}
\newcommand{\my}[1]{\textcolor{magenta}{M.-Y.L: #1}}
\newcommand{\revmy}[1]{\textcolor{magenta}{#1}}
\newcommand{\tc}[1]{\textcolor{cyan}{T.-C.W: #1}}
\newcommand{\revtc}[1]{\textcolor{cyan}{#1}}
\fi

\begin{abstract}
\vspace{-.1in}
We present a new method for synthesizing high-resolution photo-realistic images from semantic label maps using conditional generative adversarial networks (conditional GANs). Conditional GANs have enabled a variety of applications, but the results are often limited to low-resolution and still far from realistic. In this work, we generate $2048\times1024$ visually appealing results with a novel adversarial loss, as well as new multi-scale generator and discriminator architectures. Furthermore, we extend our framework to interactive visual manipulation with two additional features. First, we incorporate object instance segmentation information, which enables object manipulations such as removing/adding objects and changing the object category. Second, we propose a method to generate diverse results given the same input, allowing users to edit the object appearance interactively. Human opinion studies demonstrate that our method significantly outperforms existing methods, advancing both the quality and the resolution of deep image synthesis and editing.
\end{abstract}

\vspace{-.2in}
\section{Introduction} \lblsec{intro}

Photo-realistic image rendering using standard graphics techniques is involved, since geometry, materials, and light transport must be simulated explicitly. Although existing graphics algorithms excel at the task, building and editing virtual environments is expensive and time-consuming. That is because we have to model every aspect of the world explicitly. If we were able to render photo-realistic images using a model learned from data, we could turn the process of graphics rendering into a model learning and inference problem. Then, we could simplify the process of creating new virtual worlds by training models on new datasets. We could even make it easier to customize environments by allowing users to simply specify overall semantic structure rather than modeling geometry, materials, or lighting. 

In this paper, we discuss a new approach that produces high-resolution images from semantic label maps. This method has a wide range of applications. For example, we can use it to create synthetic training data for training visual recognition algorithms, since it is much easier to create semantic labels for desired scenarios than to generate training images. Using semantic segmentation methods, we can transform images into a semantic label domain, edit the objects in the label domain, and then transform them back to the image domain. This method also gives us new tools for higher-level image editing, e.g.,\ adding objects to images or changing the appearance of existing objects.

To synthesize images from semantic labels, one can use the \pp method, an image-to-image translation framework~\cite{isola2016image} which leverages generative adversarial networks (GANs)~\cite{goodfellow2014generative} in a conditional setting. Recently, \ck suggest that adversarial training might be unstable and prone to failure for high-resolution image generation tasks. Instead, they adopt a modified perceptual loss~\cite{gatys2016image,dosovitskiy2016generating,johnson2016perceptual} to synthesize images, which are high-resolution but often lack fine details and realistic textures.

Here we address two main issues of the above state-of-the-art methods: (1) the difficulty of generating high-resolution images with GANs~\cite{isola2016image} and (2) the lack of details and realistic textures in the previous high-resolution results~\cite{chen2017photographic}. We show that through a new, robust adversarial learning objective together with new multi-scale generator and discriminator architectures, we can synthesize photo-realistic images at $2048\times 1024$ resolution, which are more visually appealing than those computed by previous methods~\cite{isola2016image,chen2017photographic}. 
We first obtain our results with adversarial training only, without relying on any hand-crafted losses~\cite{rudin1992nonlinear} or pre-trained networks (e.g.\ VGGNet~\cite{simonyan2014very}) for perceptual losses~\cite{dosovitskiy2016generating,johnson2016perceptual} (Figs.~\ref{fig:qual_cityscape_both}c,~\ref{fig:qual_cityscape}b). Then we show that adding perceptual losses from pre-trained networks~\cite{simonyan2014very} can slightly improve the results in some circumstances (Figs.~\ref{fig:qual_cityscape_both}d,~\ref{fig:qual_cityscape}c), if a pre-trained network is available. Both results outperform previous works substantially in terms of image quality.

Furthermore, to support interactive semantic manipulation, we extend our method in two directions. First, we use instance-level object segmentation information, which can separate different object instances within the same category.
This enables flexible object manipulations, such as adding/removing objects and changing object types. Second, we propose a method to generate diverse results given the same input label map, allowing the user to edit the appearance of the same object interactively.

We compare against state-of-the-art visual synthesis systems~\cite{chen2017photographic,isola2016image}, and show that our method outperforms these approaches regarding both quantitative evaluations and human perception studies. We also perform an ablation study regarding the training objectives and the importance of instance-level segmentation information. 
In addition to semantic manipulation, we test our method on edge2photo applications (Figs.~\ref{fig:edge2photo},\ref{fig:edge2face}), which shows the generalizability of our approach.
Code and data are available at our \href{https://tcwang0509.github.io/pix2pixHD/}{website}

\begin{figure}[t]
    \centering
    \vspace{-.25in}
    \includegraphics[width=\linewidth]{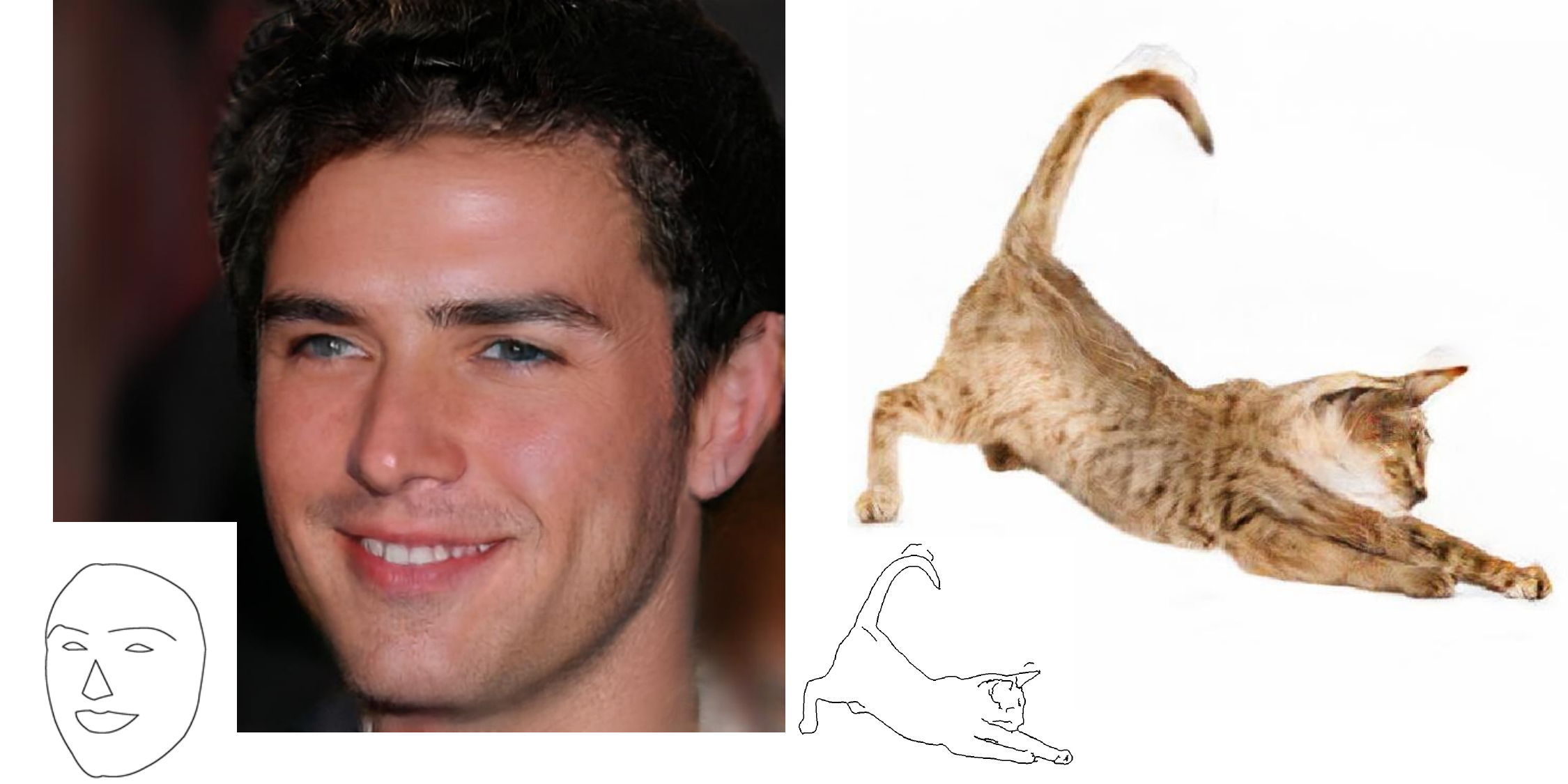}
    \vspace{-.25in}
    \caption{\small Example results of using our framework for translating edges to high-resolution natural photos, using CelebA-HQ~\cite{karras2017progressive} and internet cat images.}
    \lblfig{edge2photo}
    \vspace{-.25in}
\end{figure}.

\section{Related Work} \lblsec{related}

\paragraph{Generative adversarial networks} 
Generative adversarial networks (GANs)~\cite{goodfellow2014generative} aim to model the natural image distribution by forcing the generated samples to be indistinguishable from natural images. GANs enable a wide variety of applications such as image generation~\cite{radford2015unsupervised,zhao2016energy,arjovsky2017wasserstein}, representation learning~\cite{salimans2016improved}, image manipulation~\cite{zhu2016generative}, object detection~\cite{li2017perceptual}, and video applications~\cite{mathieu2015deep,vondrick2016generating,tulyakov2017mocogan}. Various coarse-to-fine schemes~\cite{burt1983laplacian} have been proposed~\cite{denton2015deep,huang2017sgan,zhang2016stackgan,karras2017progressive} to synthesize larger images (e.g. $256\times 256$) in an unconditional setting. Inspired by their successes, we propose a new coarse-to-fine generator and multi-scale discriminator architectures suitable for conditional image generation at a much higher resolution.

\paragraph{Image-to-image translation}
Many researchers have leveraged adversarial learning for image-to-image translation~\cite{isola2016image}, whose goal is to translate an input image from one domain to another domain given input-output image pairs as training data.
Compared to \LOneLoss loss, which often leads to blurry images~\cite{isola2016image,johnson2016perceptual}, 
the adversarial loss~\cite{goodfellow2014generative} has become a popular choice for many image-to-image tasks~\cite{wang2016generative,pathak2016context,karacan2016learning,zhang2017age,ledig2016photo,sangkloy2016scribbler,kaneko2017generative,dong2017semantic,zhu2017toward}. The reason is that the discriminator can learn a trainable loss function and automatically adapt to the differences between the generated and real images in the target domain. For example, the recent \pp framework~\cite{isola2016image} used image-conditional GANs~\cite{mirza2014conditional} for different applications, such as transforming Google maps to satellite views and generating cats from user sketches. Various methods have also been proposed to learn an image-to-image translation in the absence of training pairs~\cite{liu2016coupled,bousmalis2016unsupervised,liu2016unsupervised,shrivastava2016learning,taigman2016unsupervised,yi2017dualgan,zhu2017unpaired,tungadversarial}.

Recently, \ck suggest that it might be hard for conditional GANs to generate high-resolution images due to the training instability and optimization issues. To avoid this difficulty, they use a direct regression objective based on a perceptual loss~\cite{gatys2016image,dosovitskiy2016generating,johnson2016perceptual} and produce the first model that can synthesize $2048\times 1024$ images. The generated results are high-resolution but often lack fine details and realistic textures. Motivated by their success, we show that using our new objective function as well as novel multi-scale generators and discriminators, we not only largely stabilize the training of conditional GANs on high-resolution images, but also achieve significantly better results compared to \ck. Side-by-side comparisons clearly show our advantage (Figs.~\ref{fig:teaser},~\ref{fig:qual_cityscape_both},~\ref{fig:qual_nyu},~\ref{fig:qual_cityscape}). 

\paragraph{Deep visual manipulation}
Recently, deep neural networks have obtained promising results in various image processing tasks, such as style transfer~\cite{gatys2016image}, inpainting~\cite{pathak2016context}, colorization~\cite{zhang2016colorful}, and restoration~\cite{gharbi2016deep}. However, most of these works lack an interface for users to adjust the current result or explore the output space. 
To address this issue, Zhu~\etal~\cite{zhu2016generative} developed an optimization method for editing the object appearance based on the priors learned by GANs. Recent works~\cite{isola2016image,sangkloy2016scribbler,zhang2017real} also provide user interfaces for creating novel imagery from low-level cues such as color and sketch. All of the prior works report results on low-resolution images. Our system shares the same spirit as this past work, but we focus on object-level semantic editing, allowing users to interact with the entire scene and manipulate individual objects in the image. As a result, users can quickly create a new scene with minimal effort. Our interface is inspired by prior data-driven graphics systems~\cite{johnson2006semantic,lalonde2007photo,chen2009sketch2photo}. But our system allows more flexible manipulations and produces high-res results in real-time. 
\section{Instance-Level Image Synthesis}
\lblsec{alg}
We propose a conditional adversarial framework for generating high-resolution photo-realistic images from semantic label maps. We first review our baseline model \pp (\refsec{alg:pix2pix}). We then describe how we increase the photo-realism and resolution of the results with our improved objective function and network design (\refsec{alg:resolution}). Next, we use additional instance-level object semantic information to further improve the image quality (\refsec{alg:inst}). Finally, we introduce an instance-level feature embedding scheme to better handle the multi-modal nature of image synthesis, which enables interactive object editing (\refsec{alg:feat}).

\subsection{The pix2pix Baseline} \lblsec{alg:pix2pix}
The \pp method~\cite{isola2016image} is a conditional GAN framework for image-to-image translation. It consists of a generator $G$ and a discriminator $D$. For our task, the objective of the generator $G$ is to translate semantic label maps to realistic-looking images, while the discriminator $D$ aims to distinguish real images from the translated ones. The framework operates in a supervised setting. In other words, the training dataset is given as a set of pairs of corresponding images $\{(\mathbf{s_i},\mathbf{x_i})\}$, where $\mathbf{s_i}$ is a semantic label map and $\mathbf{x_i}$ is a corresponding natural photo. Conditional GANs aim to model the conditional distribution of real images given the input semantic label maps via the following minimax game: 
\begin{equation}
\min_{G} \max_{D} \mathcal{L}_{\text{GAN}}(G,D)
\label{eqn::minimax}
\end{equation}
where the objective function $\mathcal{L}_{GAN}(G,D)$
\footnote{we denote $\mathbb{E}_{\s} \triangleq \mathbb{E}_{\s\sim p_{\text{data}}(s)}$ and $ \mathbb{E}_{(\s,\x)} \triangleq \mathbb{E}_{(\s,\x) \sim p_{\text{data}}(\s, \x)}$ for simplicity.}
 is given by 
\begin{equation}
\mathbb{E}_{(\s,\x)}[\log D(\s,\x)]+ 
\mathbb{E}_{\s}[\log (1- D(\s,G(\s))].
\label{eqn::gan_problem}
\end{equation}
The \pp method adopts U-Net~\cite{ronneberger2015u} as the generator and a patch-based fully convolutional network~\cite{long2015fully} as the discriminator. The input to the discriminator is a channel-wise concatenation of the semantic label map and the corresponding image. 
However, the resolution of the generated images on Cityscapes~\cite{Cordts2016cityscapes} is up to $256\times 256$. We tested directly applying the \pp framework to generate high-resolution images but found the training unstable and the quality of generated images unsatisfactory. Therefore, we describe how we improve the \pp framework in the next subsection.

\subsection{Improving Photorealism and Resolution} \lblsec{alg:resolution}
We improve the \pp framework by using a coarse-to-fine generator, a multi-scale discriminator architecture, and a robust adversarial learning objective function.

\begin{figure*}[ht!]
  \centering
  \includegraphics[width=0.95\linewidth]{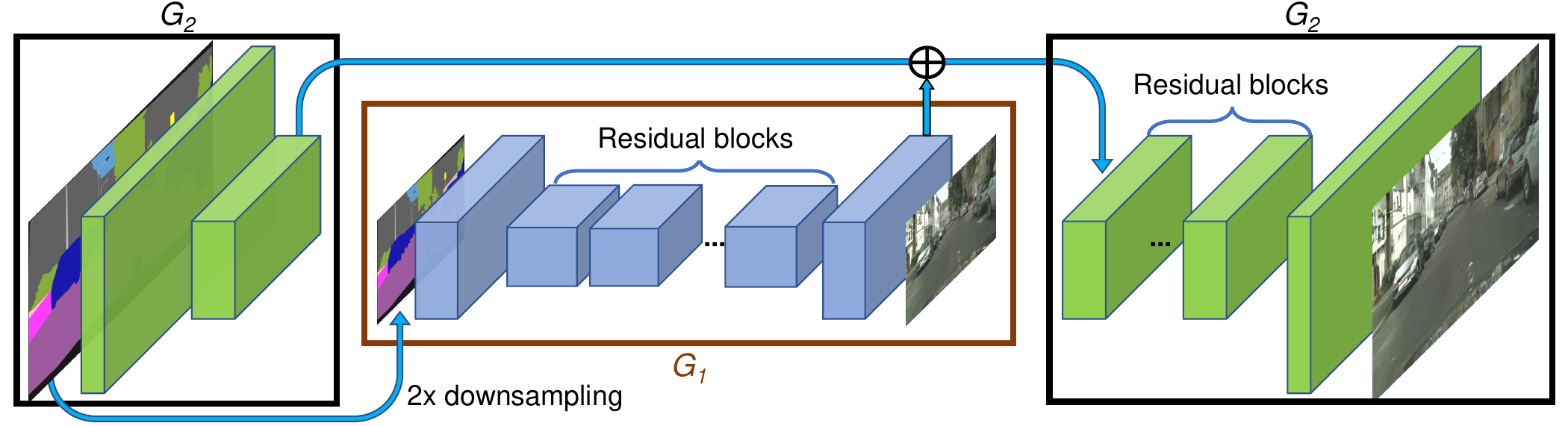}   
  \caption{Network architecture of our generator. We first train a residual network $G_1$ on lower resolution images. Then, another residual network $G_2$ is appended to $G_1$ and the two networks are trained jointly on high resolution images. Specifically, the input to the residual blocks in $G_2$ is the element-wise sum of the feature map from $G_2$ and the last feature map from $G_1$.}
  \lblfig{arch}
\end{figure*}

\noindent {\bf Coarse-to-fine generator} We decompose the generator into two sub-networks: $G_1$ and $G_2$. We term $G_1$ as the global generator network and $G_2$ as the local enhancer network. The generator is then given by the tuple  $G=\{G_1,G_2\}$ as visualized in \reffig{arch}. The global generator network operates at a resolution of $1024\times 512$, and the local enhancer network outputs an image with a resolution that is $4\times$ the output size of the previous one ($2\times$ along each image dimension). For synthesizing images at an even higher resolution, additional local enhancer networks could be utilized. For example, the output image resolution of the generator $G=\{G_1, G_2\}$ is $2048\times 1024$, and the output image resolution of $G=\{G_1, G_2, G_3\}$ is $4096\times 2048$.

Our global generator is built on the architecture proposed by Johnson \etal~\cite{johnson2016perceptual}, which has been proven successful for neural style transfer on images up to $512\times 512$. It consists of $3$ components: a convolutional front-end $G_1^{(F)}$, a set of residual blocks $G_1^{(R)}$~\cite{he2016deep}, and a transposed convolutional back-end $G_1^{(B)}$. A semantic label map of resolution $1024\times 512$ is passed through the 3 components sequentially to output an image of resolution $1024\times 512$.

The local enhancer network also consists of 3 components: a convolutional front-end $G_2^{(F)}$, a set of residual blocks $G_2^{(R)}$, and a transposed convolutional back-end $G_2^{(B)}$. The resolution of the input label map to $G_2$ is $2048\times 1024$. Different from the global generator network, the input to the residual block $G_2^{(R)}$ is the element-wise sum of two feature maps: the output feature map of $G_2^{(F)}$, and the last feature map of the back-end of the global generator network $G_1^{(B)}$. This helps integrating the global information from $G_1$ to $G_2$. 

During training, we first train the global generator and then train the local enhancer in the order of their resolutions. We then jointly fine-tune all the networks together. We use this generator design to effectively aggregate global and local information for the image synthesis task. We note that such a multi-resolution pipeline is a well-established practice in computer vision~\cite{burt1983laplacian} and two-scale is often enough~\cite{brown2003recognising}. Similar ideas but different architectures could be found in recent unconditional GANs~\cite{denton2015deep,huang2017sgan} and conditional image generation~\cite{chen2017photographic,zhang2016stackgan}.

\noindent {\bf Multi-scale discriminators } High-resolution image synthesis poses a significant challenge to the GAN discriminator design. To differentiate high-resolution real and synthesized images, the discriminator needs to have a large receptive field. This would require either a deeper network or larger convolutional kernels, both of which would increase the network capacity and potentially cause overfitting. Also, both choices demand a larger memory footprint for training, which is already a scarce resource for high-resolution image generation.

To address the issue, we propose using multi-scale discriminators. We use $3$ discriminators that have an identical network structure but operate at different image scales. We will refer to the discriminators as $D_1$, $D_2$ and $D_3$. Specifically, we downsample the real and synthesized high-resolution images by a factor of $2$ and $4$ to create an image pyramid of 3 scales. The discriminators $D_1$, $D_2$ and $D_3$ are then trained to differentiate real and synthesized images at the $3$ different scales, respectively. Although the discriminators have an identical architecture, the one that operates at the coarsest scale has the largest receptive field. It has a more global view of the image and can guide the generator to generate globally consistent images. On the other hand, the discriminator at the finest scale encourages the generator to produce finer details. This also makes training the coarse-to-fine generator easier, since extending a low-resolution model to a higher resolution only requires adding a discriminator at the finest level, rather than retraining from scratch.
Without the multi-scale discriminators, we observe that many repeated patterns often appear in the generated images.

With the discriminators, the learning problem in Eq.~(\ref{eqn::minimax}) then becomes a multi-task learning problem of 
\begin{equation}
\small
\min_{G} \max_{D_1,D_2,D_3} \sum_{k=1,2,3}\mathcal{L}_{\text{GAN}}(G,D_k).
\end{equation}
Using multiple GAN discriminators at the same image scale has been proposed in unconditional GANs~\cite{durugkar2016generative}. Iizuka et al.~\cite{iizuka2017globally} add a global image classifier to conditional GANs to synthesize globally coherent content for inpainting.
Here we extend the design to multiple discriminators at different image scales for modeling high-resolution images. 

\noindent {\bf Improved adversarial loss } We improve the GAN loss in Eq.~(\ref{eqn::gan_problem}) by incorporating a feature matching loss based on the discriminator. This loss stabilizes the training as the generator has to produce natural statistics at multiple scales. Specifically, we extract features from multiple layers of the discriminator and learn to match these intermediate representations from the real and the synthesized image. For ease of presentation, we denote the $i$th-layer feature extractor of discriminator $D_k$ as $D_k^{(i)}$ (from input to the $i$th layer of $D_k$). The feature matching loss $\mathcal{L}_{\text{FM}}(G,D_k)$ is then calculated as:
\begin{equation}
\small
    \mathcal{L}_{\text{FM}}(G,D_k) = \mathbb{E}_{(\s,\x)} \sum_{i=1}^T \frac{1}{N_i}[||D_k^{(i)}(\s,\x)-D_k^{(i)}(\s,G(\s))||_1],
\end{equation}
where $T$ is the total number of layers and $N_i$ denotes the number of elements in each layer.  Our GAN discriminator feature matching loss is related to the perceptual loss~\cite{gatys2016image,johnson2016perceptual,dosovitskiy2016generating}, which has been shown to be useful for image super-resolution~\cite{ledig2016photo} and style transfer~\cite{johnson2016perceptual}. 
In our experiments, we discuss how the discriminator feature matching loss and the perceptual loss can be jointly used for further improving the performance. 
We note that a similar loss is used in VAE-GANs~\cite{larsen2015autoencoding}.

Our full objective combines both GAN loss and feature matching loss as:
\begin{equation}
\footnotesize
\min_{G}\bigg(\Big(\max_{D_1,D_2,D_3} \sum_{k=1,2,3}\mathcal{L}_{\text{GAN}}(G,D_k)\Big) + \lambda \sum_{k=1,2,3}\mathcal{L}_{\text{FM}}(G,D_k)\bigg)
\label{eqn::multigan}
\end{equation}
where $\lambda$ controls the importance of the two terms. Note that for the feature matching loss $\mathcal{L}_{\text{FM}}$, $D_k$ only serves as a feature extractor and does not maximize the loss $\mathcal{L}_{\text{FM}}$.

\subsection{Using Instance Maps} \lblsec{alg:inst}
\begin{figure}
  \centering
  \includegraphics[width=1.0\linewidth]{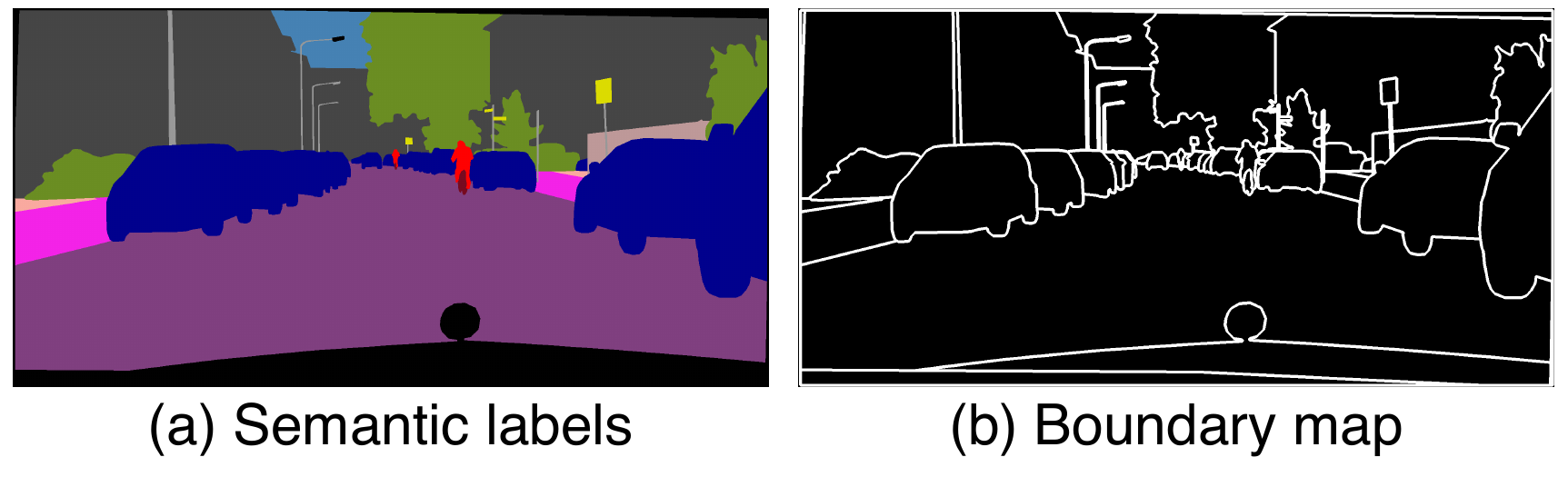}   
  \caption{Using instance maps: (a) a typical semantic label map. Note that all connected cars have the same label, which makes it hard to tell them apart.
  (b) The extracted instance boundary map. With this information, separating different objects becomes much easier.
  }  
  \lblfig{inst}
\end{figure}

\begin{figure}
  \centering
  \includegraphics[width=1.0\linewidth]{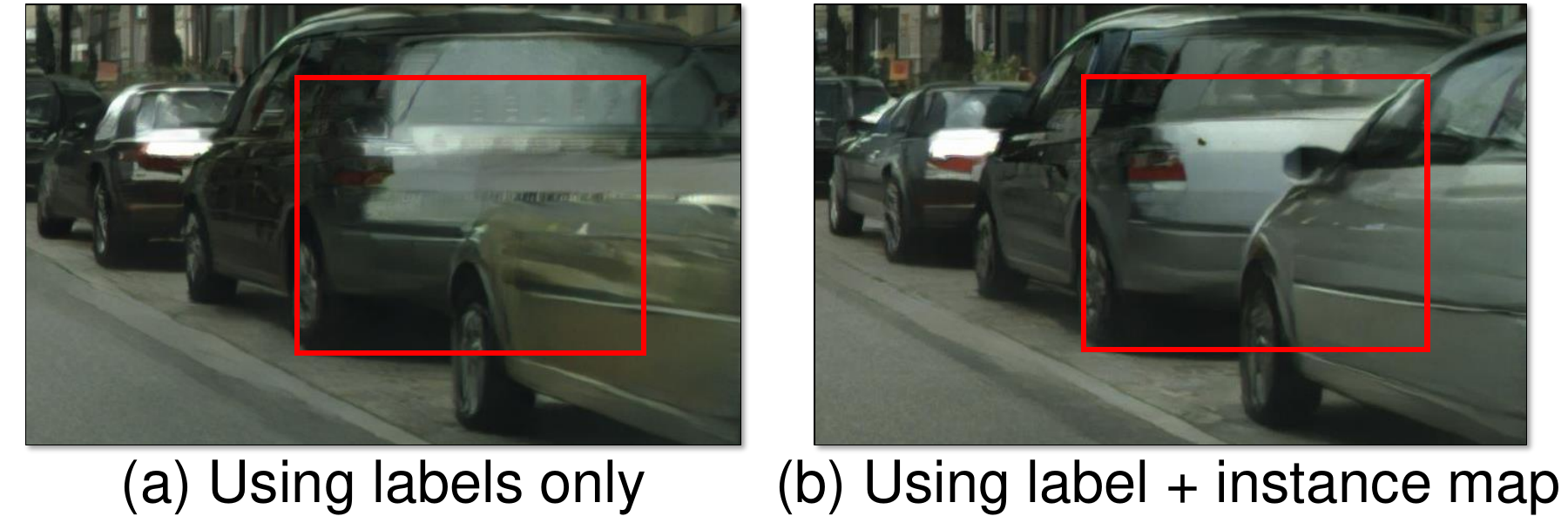}   
  \caption{Comparison between results without and with instance maps. It can be seen that when instance boundary information is added, adjacent cars have sharper boundaries.}
  \lblfig{inst_compare}
\end{figure}

Existing image synthesis methods only utilize semantic label maps~\cite{isola2016image,karacan2016learning,chen2017photographic}, an image where each pixel value represents the object class of the pixel. This map does not differentiate objects of the same category. On the other hand, an instance-level semantic label map contains a unique object ID for each individual object. 
To incorporate the instance map, one can directly pass it into the network, or encode it into a one-hot vector. However, both approaches are difficult to implement in practice, since different images may contain different numbers of objects of the same category. Alternatively, one can pre-allocate a fixed number of channels (e.g., $10$) for each class, but this method fails when the number is set too small, and wastes memory when the number is too large.

Instead, we argue that the most critical information the instance map provides, which is not available in the semantic label map, is the object boundary. 
For example, when objects of the same class are next to one another, looking at the semantic label map alone cannot tell them apart. This is especially true for the street scene since many parked cars or walking pedestrians are often next to one another, as shown in \reffig{inst}a. However, with the instance map, separating these objects becomes an easier task.

Therefore, to extract this information, we first compute the instance boundary map (\reffig{inst}b). In our implementation, a pixel in the instance boundary map is $1$ if its object ID is different from any of its $4$-neighbors, and $0$ otherwise. 
The instance boundary map is then concatenated with the one-hot vector representation of the semantic label map, and fed into the generator network. 
Similarly, the input to the discriminator is the channel-wise concatenation of instance boundary map, semantic label map, and the real/synthesized image. Figure~\ref{fig:inst_compare}b shows an example demonstrating the improvement by using object boundaries. 
Our user study in \refsec{results} also shows the model trained with instance boundary maps renders more photo-realistic object boundaries.

\subsection{Learning an Instance-level Feature Embedding}
\lblsec{alg:feat}
\begin{figure}
  \centering
  \includegraphics[width=.95\linewidth]{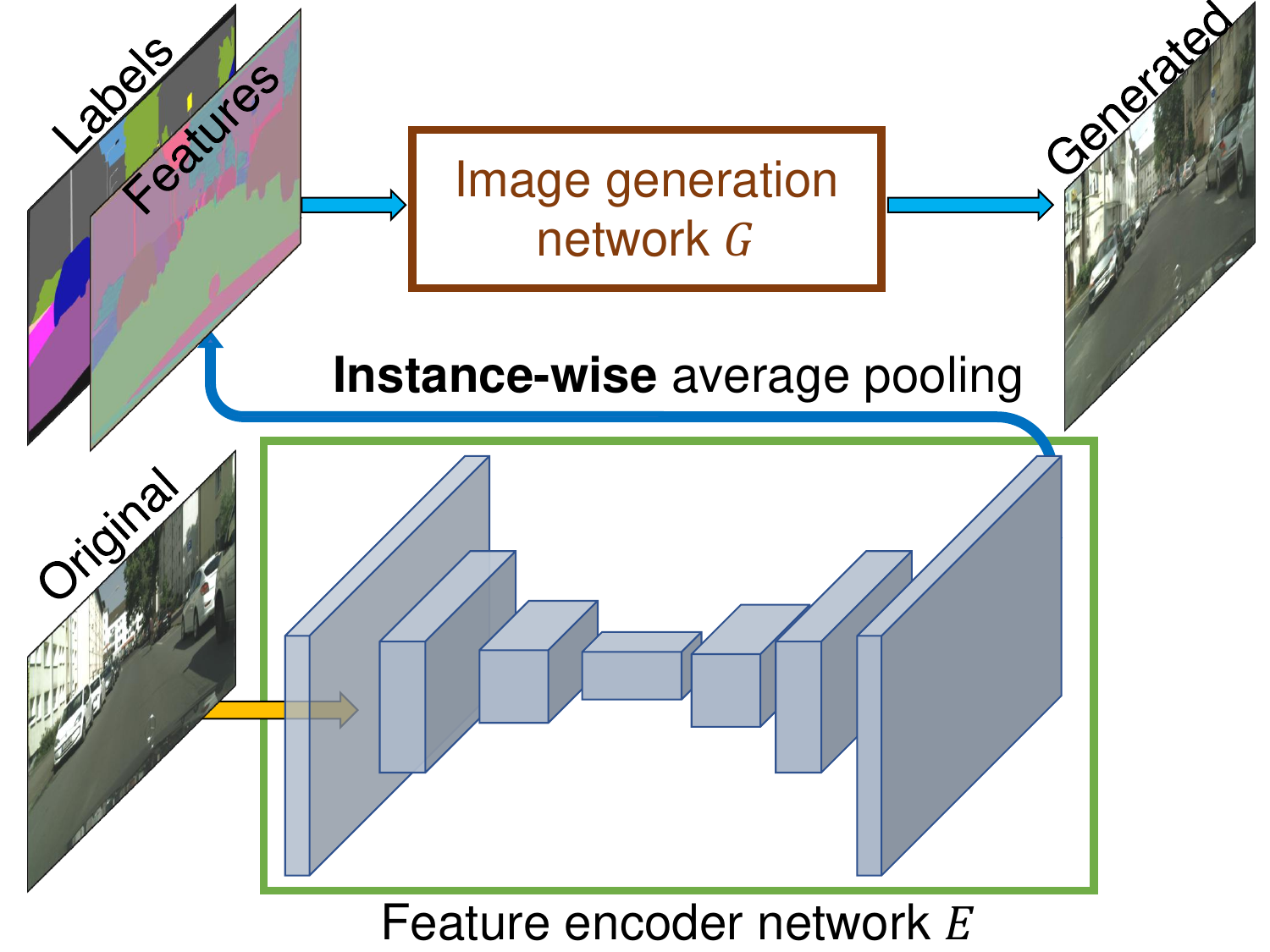}   
  \caption{Using instance-wise features in addition to labels for generating images. 
  }
  \lblfig{feat}
\end{figure}

Image synthesis from semantic label maps is a one-to-many mapping problem. An ideal image synthesis algorithm should be able to generate diverse, realistic images using the same semantic label map. Recently, several works learn to produce a fixed number of discrete outputs given the same input~\cite{ghosh2017multi,chen2017photographic} or synthesize diverse modes controlled by a latent code that encodes the entire image~\cite{zhu2017toward}. Although these approaches tackle the multi-modal image synthesis problem, they are unsuitable for our image manipulation task mainly for two reasons.
First, the user has no intuitive control over which kinds of images the model would produce\cite{ghosh2017multi,chen2017photographic}.
Second, these methods focus on global color and texture changes and allow no object-level control on the generated contents. 

To generate diverse images and allow instance-level control, we propose adding additional low-dimensional feature channels as the input to the generator network. We show that, by manipulating these features, we can have flexible control over the image synthesis process. Furthermore, note that since the feature channels are continuous quantities, our model is, in principle, capable of generating infinitely many images.

To generate the low-dimensional features, we train an encoder network $E$ to find a low-dimensional feature vector that corresponds to the ground truth target for each instance in the image. Our feature encoder architecture is a standard encoder-decoder network. To ensure the features are consistent within each instance, we add an instance-wise average pooling layer to the output of the encoder to compute the average feature for the object instance. The average feature is then broadcast to all the pixel locations of the instance. Figure~\ref{fig:feat} visualizes an example of the encoded features.

We replace $G(\s)$ with $G(\s, E(\x))$ in Eq.~(\ref{eqn::multigan}) and train the encoder jointly with the generators and discriminators. After the encoder is trained, we run it on all instances in the training images and record the obtained features. Then we perform a $K$-means clustering on these features for each semantic category.
Each cluster thus encodes the features for a specific style, for example, the asphalt or cobblestone texture for a road.
At inference time, we randomly pick one of the cluster centers and use it as the encoded features. These features are concatenated with the label map and used as the input to our generator. We tried to enforce the Kullback-Leibler loss~\cite{kingma2013auto} on the feature space for better test-time sampling as used in the recent work~\cite{zhu2017toward} but found it quite involved for users to adjust the latent vectors for each object directly. Instead, for each object instance, we  present $K$ modes for users to choose from.

\section{Results} \lblsec{results}

We first provide a quantitative comparison against leading methods in \refsec{results:quan}. We then report a subjective human perceptual study in \refsec{results:qual}. Finally, we show a few examples of interactive object editing results in \refsec{results:ui}. 

\vspace{.03in}
{\noindent \bf Implementation details}
We use LSGANs~\cite{mao2017least} for stable training. In all experiments, we set the weight ${\lambda=10}$ (Eq.~(\ref{eqn::multigan})) and ${K=10}$ for K-means. We use $3$-dimensional vectors to encode features for each object instance. We experimented with adding a perceptual loss $\lambda \sum_{i=1}^N\frac{1}{M_i}[||F^{(i)}(\x)-F^{(i)}(G(\s))||_1]$ to our objective (Eq.~(\ref{eqn::multigan})), where $\lambda=10$ and $F^{(i)}$ denotes the $i$-th layer with $M_i$  elements of the VGG network. We observe that this loss slightly improves the results. We name these two variants as \textbf{\ours} and \textbf{ours (w/o VGG loss)}. Please find more training and architecture details in the appendix.

\vspace{.03in}
{\noindent \bf Datasets} We conduct extensive comparisons and ablation studies on Cityscapes dataset~\cite{Cordts2016cityscapes} and NYU Indoor RGBD dataset~\cite{Silberman2012indoor}. We report additional qualitative results on ADE20K dataset~\cite{zhou2017scene} and Helen Face dataset~\cite{le2012interactive,smith2013exemplar}.

\vspace{.03in}
{\noindent \bf Baselines} We compare our method with two state-of-the-art algorithms: \pp~\cite{isola2016image} and CRN~\cite{chen2017photographic}. We train \pp models on high-res images with the default setting. 
We produce the high-res CRN images via the authors' publicly available model.

\subsection{Quantitative Comparisons} \lblsec{results:quan}

We adopt the same evaluation protocol from previous image-to-image translation works~\cite{isola2016image,zhu2017unpaired}.
To quantify the quality of our results, we perform semantic segmentation on the synthesized images and compare how well the predicted segments match the input. The intuition is that if we can produce realistic images that correspond to the input label map, an off-the-shelf semantic segmentation model (e.g.,\ PSPNet~\cite{zhao2017pspnet} that we use) should be able to predict the ground truth label. \reftbl{seg} reports 
the calculated segmentation accuracy.  
As can be seen, for both pixel-wise accuracy and mean intersection-over-union (IoU), our method outperforms the other methods by a large margin.
Moreover, our result is very close to the result of the original images, the theoretical ``upper bound'' of the realism we can achieve.
This justifies the superiority of our algorithm.

\begin{table}[t!]
\centering
\begin{tabular}{  c  c  c  c | c  }
\toprule
\multicolumn{1}{c}{}
 & \pp~\cite{isola2016image} & CRN~\cite{chen2017photographic} & Ours & Oracle \\ \midrule 
Pixel acc & 78.34 &  70.55  &  {\bf 83.78}  &  84.29 \\ 
Mean IoU & 0.3948  &  0.3483  &  {\bf 0.6389}  &  0.6857 \\ \bottomrule 
\end{tabular}
\caption{Semantic segmentation scores on results by different methods on the Cityscapes dataset~\cite{Cordts2016cityscapes}. Our result outperforms the other methods by a large margin and is very close to the accuracy on original images (i.e., the oracle).}
\lbltbl{seg}
\end{table}

\subsection{Human Perceptual Study} \lblsec{results:qual}
We further evaluate our algorithm via a human subjective study.
We perform pairwise A/B tests deployed on the Amazon Mechanical Turk (MTurk) platform on the Cityscapes dataset~\cite{Cordts2016cityscapes}.
We follow the same experimental procedure as described in \ck.
More specifically, two different kinds of experiments are conducted: unlimited time and limited time, as explained below.

\vspace{.03in}
{\noindent \bf Unlimited time} For this task, workers are given two images at once, each of which is synthesized by a different method for the same label map. We give them unlimited time to select which image looks more natural.
The left-right order and the image order are randomized to ensure fair comparisons. All $500$ Cityscapes test images are compared $10$ times, resulting in $5,000$ human judgments for each method. 
In this experiment, we use the model trained on labels only (without instance maps) to ensure a fair comparison.
\reftbl{unlimited} shows that both variants of our method outperform the other methods significantly.

\begin{table}[t!]
\setlength{\tabcolsep}{2pt}
\centering
\begin{tabularx}{0.45\textwidth}{c *{2}{Y}} \toprule
\multicolumn{1}{c}{}
& \pp~\cite{isola2016image} & CRN~\cite{chen2017photographic}  \\ \midrule
Ours & 93.8\% & 86.2\%  \\ 
Ours {\footnotesize (w/o VGG)}  & 94.6\%  &  85.2\% \\ \bottomrule
\end{tabularx}
\caption{Pairwise comparison results on the Cityscapes dataset~\cite{Cordts2016cityscapes} (unlimited time). Each cell lists the percentage where our result is preferred over the other method. Chance is at $50\%$.}
\lbltbl{unlimited}
\end{table}

\vspace{.03in}
{\noindent \bf Limited time} Next, for the limited time experiment, we compare our result with CRN and the original image (ground truth).
In each comparison, we show results of two methods for a short period of time. 
We randomly select a duration between $1/8$ seconds and $8$ seconds, as adopted by prior work~\cite{chen2017photographic}.
This evaluates how quickly the difference between the images can be perceived. 
\reffig{limited} shows the comparison results at different time intervals. 
As the given time becomes longer and longer, the differences between these three types of images become more apparent and easier to observe.
Figures~\ref{fig:qual_cityscape_both} and \ref{fig:qual_cityscape} show some example results. 

\begin{figure}
  \centering
  \includegraphics[width=.95\linewidth]{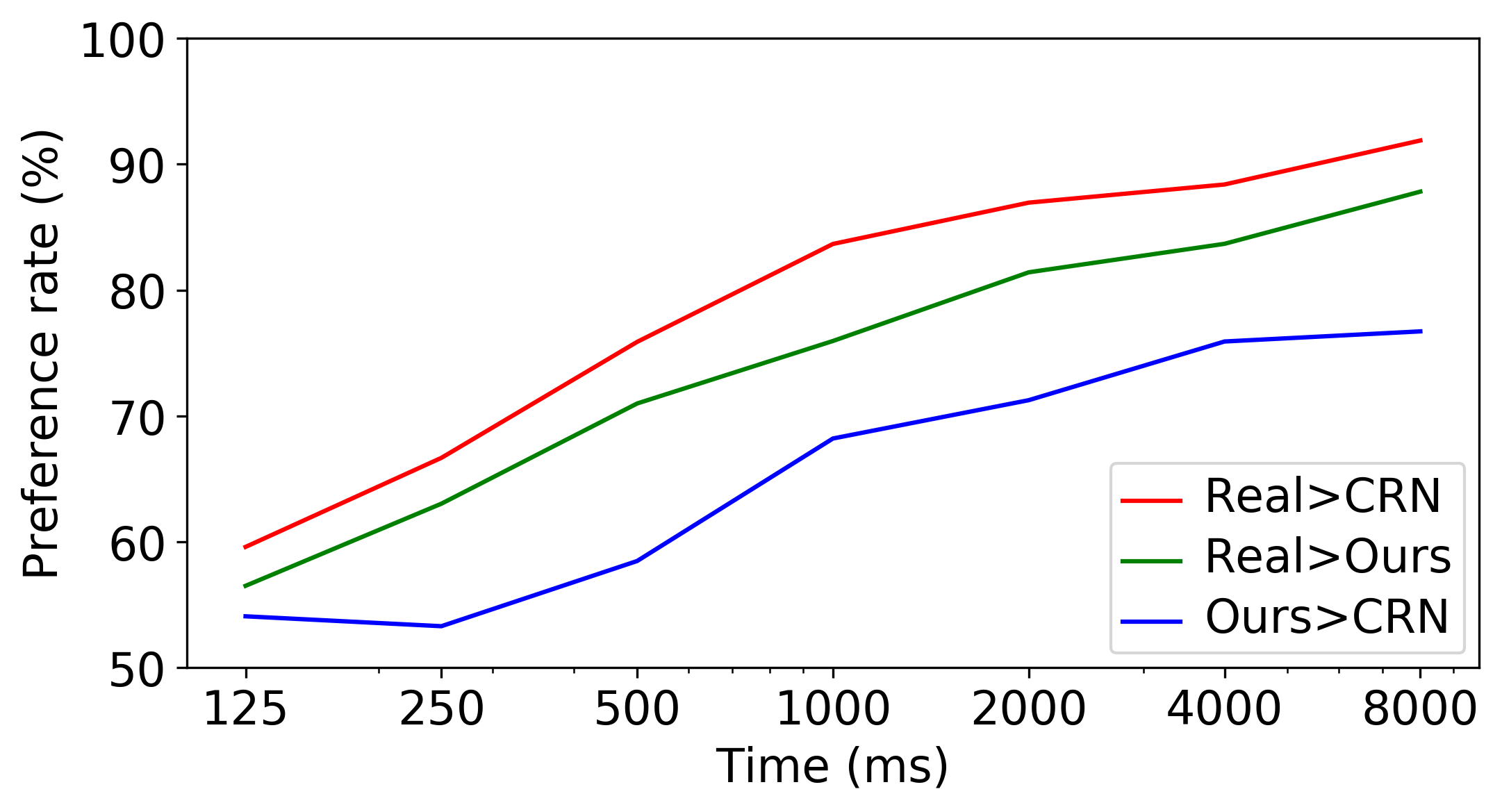} 
  \caption{Limited time comparison results. Each line shows the percentage when one method is preferred over the other.}
  \lblfig{limited}
\end{figure}

\vspace{.03in}
{\noindent \bf Analysis of the loss function} We also study the importance of each term in our objective function using the \textit{unlimited time}  experiment.
Specifically, our final loss contains three components: GAN loss, discriminator-based feature matching loss, and VGG perceptual loss. We compare our final implementation to the results using ($1$) only GAN loss, and ($2$) GAN $+$ feature matching loss (i.e.,\ without VGG loss). The obtained preference rates are $68.55\%$ and $58.90\%$, respectively.
As can be seen, adding the feature matching loss substantially improves the performance, while adding perceptual loss further enhances the results. However, note that using the perceptual loss is not critical, and we are still able to generate visually appealing results even without it (e.g.,\ Figs.~\ref{fig:qual_cityscape_both}c,~\ref{fig:qual_cityscape}b). 

\vspace{.03in}
{\noindent \bf Using instance maps} We compare the results using instance maps to results without using them. We highlight the car regions in the images and ask the participants to choose which region looks more realistic. We obtain a preference rate of $64.34\%$, which indicates that using instance maps improves the realism of our results, especially around the object boundaries.

\vspace{.03in}
{\noindent \bf Analysis of the generator } 
We compare results of different generators with all the other components fixed. In particular, we compare our generator with two state-of-the-art generator architectures: U-Net~\cite{ronneberger2015u,isola2016image} and CRN~\cite{chen2017photographic}. We evaluate the performance regarding both semantic segmentation scores and human perceptual study results. \reftbl{seg_G} and \reftbl{unlimited_G} show that our coarse-to-fine generator outperforms other networks by a large margin.

\vspace{.03in}
{\noindent \bf Analysis of the discriminator } 
Next, we also compare results using our multi-scale discriminators and results using only one discriminator while we keep the generator and the loss function fixed. The segmentation scores on Cityscapes~\cite{Cordts2016cityscapes} (\reftbl{seg_D}) demonstrate that using multi-scale discriminators helps produce higher quality results as well as stabilize the adversarial training. We also perform pairwise A/B tests on the Amazon Mechanical Turk platform. $69.2\%$ of the participants prefer our results with multi-scale discriminators over the results trained with a single-scale discriminator (Chance is $50\%$).

\begin{table}[t!]
\vspace{-.2in}
\centering
\begin{tabular}{  c  c  c  c  }
\toprule
\multicolumn{1}{c}{}
 & U-Net~\cite{ronneberger2015u,isola2016image} & CRN~\cite{chen2017photographic} & Our generator \\ \midrule 
Pixel acc (\%) & 77.86 &  78.96  &  {\bf 83.78} \\
Mean IoU & 0.3905  &  0.3994  &  {\bf 0.6389} \\ \bottomrule 
\end{tabular}
\caption{Semantic segmentation scores on results using different generators on the Cityscapes dataset~\cite{Cordts2016cityscapes}. Our generator obtains the highest scores.}
\lbltbl{seg_G}
\end{table}

\begin{table}[t!]
\setlength{\tabcolsep}{2pt}
\centering
\begin{tabularx}{0.45\textwidth}{c *{2}{Y}} \toprule
& U-Net~\cite{ronneberger2015u,isola2016image} & CRN~\cite{chen2017photographic}   \\ \midrule
Our generator & 80.0\% & 76.6\%  \\ 
\bottomrule 
\end{tabularx}
\caption{Pairwise comparison results on the Cityscapes dataset~\cite{Cordts2016cityscapes}. Each cell lists the percentage where our result is preferred over the other method. Chance is at $50\%$.}
\lbltbl{unlimited_G}
\end{table}

\begin{table}[h!]
\vspace{-.1in}
\centering
\begin{tabularx}{0.38\textwidth}{  c  c  c  }
\toprule
\multicolumn{1}{c}{}
 & single D & multi-scale Ds \\ \midrule 
Pixel acc (\%) & 82.87 &  {\bf 83.78} \\ 
Mean IoU & 0.5775 &  {\bf 0.6389} \\ \bottomrule 
\end{tabularx}
\caption{Semantic segmentation scores on results using either a single discriminator (\texttt{single D}) or multi-scale discriminators (\texttt{multi-scale Ds}) on the Cityscapes dataset~\cite{Cordts2016cityscapes}. Using multi-scale discriminators helps improve the segmentation scores.}
\lbltbl{seg_D}
\end{table}

{\noindent \bf Additional datasets } To further evaluate our method, we perform unlimited time comparisons on the NYU dataset. We obtain $86.7\%$ and $63.7\%$ against \pp and CRN, respectively. \reffig{qual_nyu} show some example images. 
Finally, we show results on the ADE20K~\cite{zhou2017scene} dataset (\reffig{qual_ade}).

\begin{figure}
  \vspace{-.2in}
  \centering
  \includegraphics[width=\linewidth]{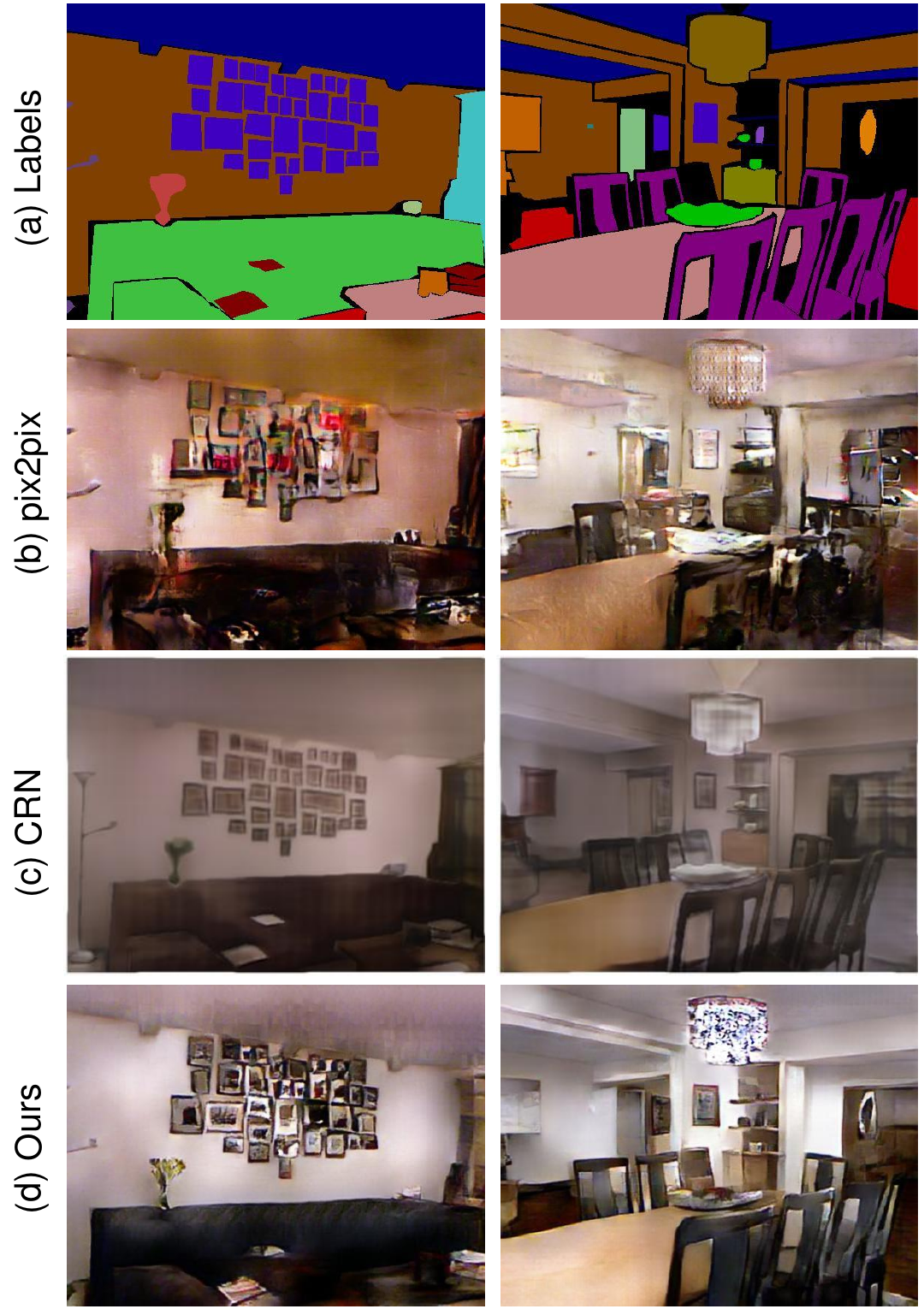}  
  \caption{Comparison on the NYU dataset~\cite{Silberman2012indoor}. Our method generates more realistic and colorful images than the other methods.}
  \lblfig{qual_nyu}
  \vspace{-.1in}
\end{figure}

\begin{figure*}
  \vspace{-.3in}
  \raggedleft
  \includegraphics[width=.95\linewidth]{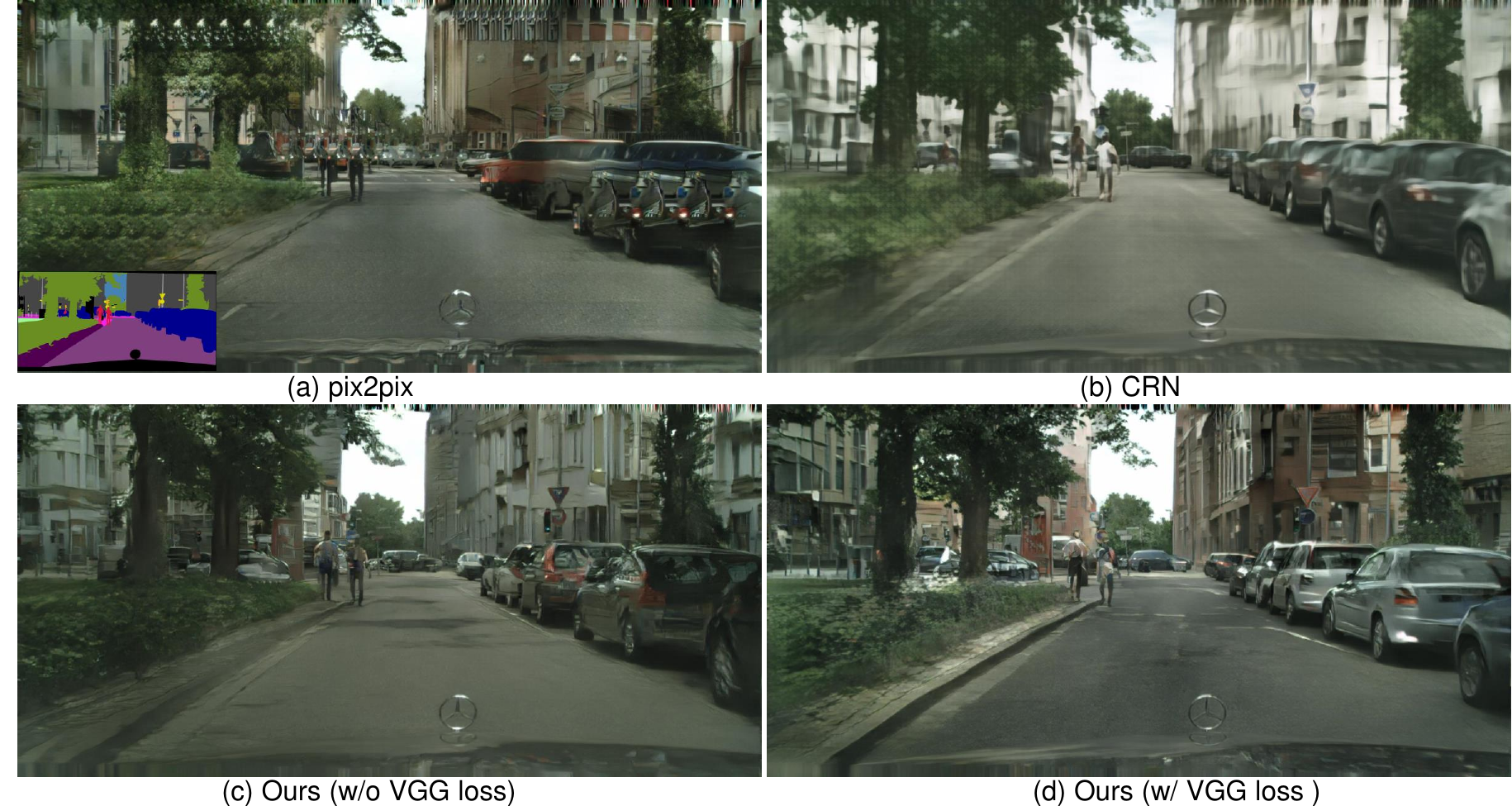}  
  \vspace{-.1in}
  \caption{Comparison on the Cityscapes dataset~\cite{Cordts2016cityscapes} (label maps shown at the lower left corner in (a)). For both without and with VGG loss, our results are more realistic than the other two methods. Please zoom in for details.}
  \lblfig{qual_cityscape_both}
\end{figure*}

\begin{figure*}
  \vspace{-.1in}
  \raggedleft
  \includegraphics[width=.97\linewidth]{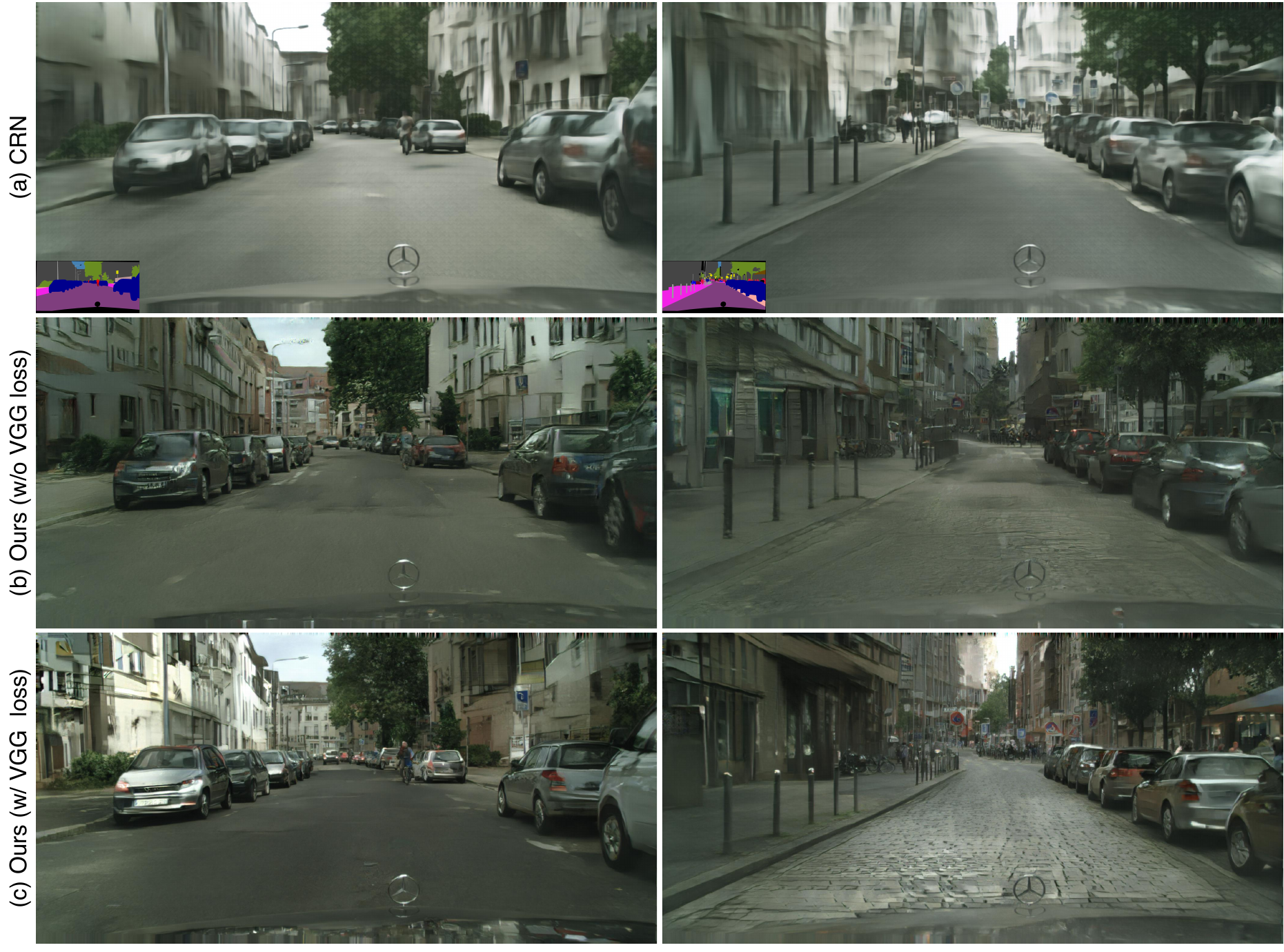}  
  \vspace{-.1in}
  \caption{Additional comparison results with CRN~\cite{chen2017photographic} on the Cityscapes dataset. Again, both our results have finer details in the synthesized cars, the trees, the buildings, etc. Please zoom in for details.}
  \lblfig{qual_cityscape}
  \vspace{-.15in}
\end{figure*}

\subsection{Interactive Object Editing} \lblsec{results:ui}
Our feature encoder allows us to perform interactive instance editing on the resulting images.
For example, we can change the object labels in the image to quickly create novel scenes, such as replacing trees with buildings (\reffig{teaser}b).
We can also change the colors of individual cars or the textures of the road (\reffig{teaser}c).
Please check out our interactive demos on our website.

Besides, we implement our interactive object editing feature on the Helen Face dataset where labels for different facial parts are available~\cite{smith2013exemplar} (\reffig{qual_face}).
This makes it easy to edit human portraits, e.g.,\ changing the face color to mimic different make-up effects or adding beard to a face.

\section{Discussion and Conclusion} \lblsec{conclusion}

\begin{figure}
  \centering
  \includegraphics[width=\linewidth]{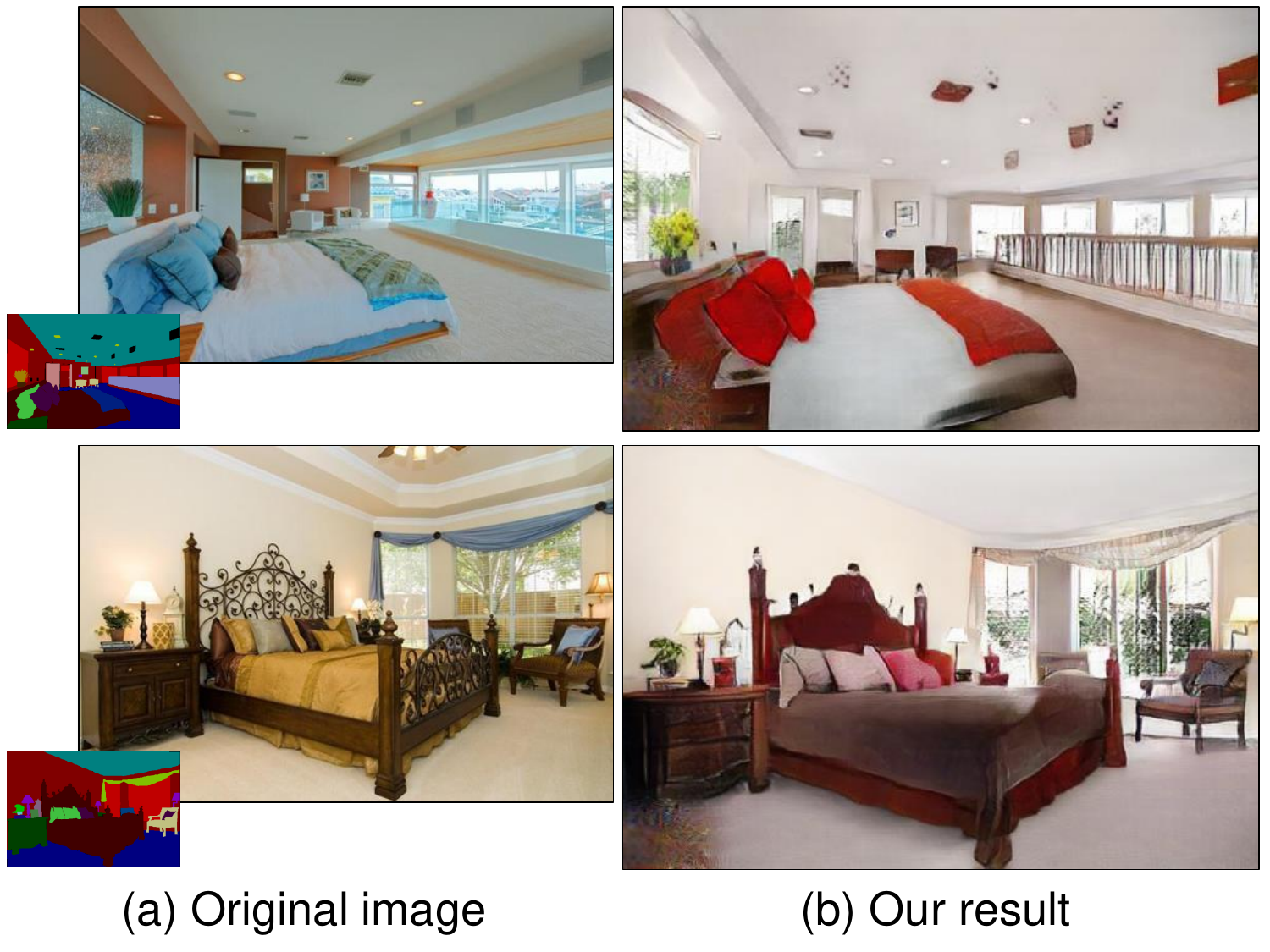}  
  \caption{Results on the ADE20K dataset~\cite{zhou2017scene} (label maps shown at lower left corners in (a)). Our method generates images at similar level of realism as the original images.}
  \lblfig{qual_ade}
\end{figure}

\begin{figure}
  \centering
  \includegraphics[width=\linewidth]{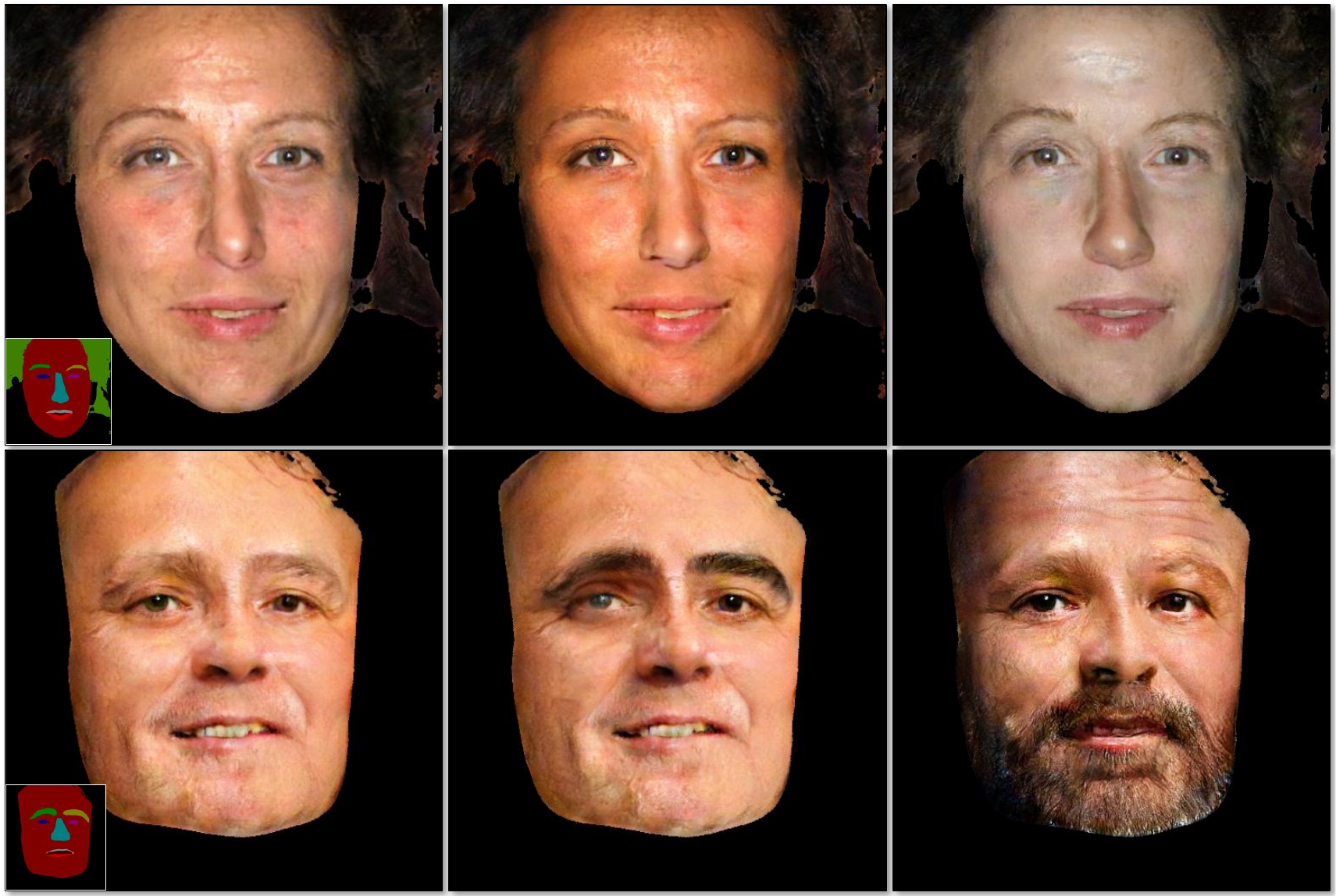}  
  \caption{Diverse results on the Helen Face dataset~\cite{smith2013exemplar} (label maps shown at lower left corners). With our interface, a user can edit the attributes of individual facial parts in real-time, such as changing the skin color or adding eyebrows and beards. See our video for more details.
  }
  \lblfig{qual_face}
\end{figure}

\begin{figure}
  \centering
  \includegraphics[width=\linewidth]{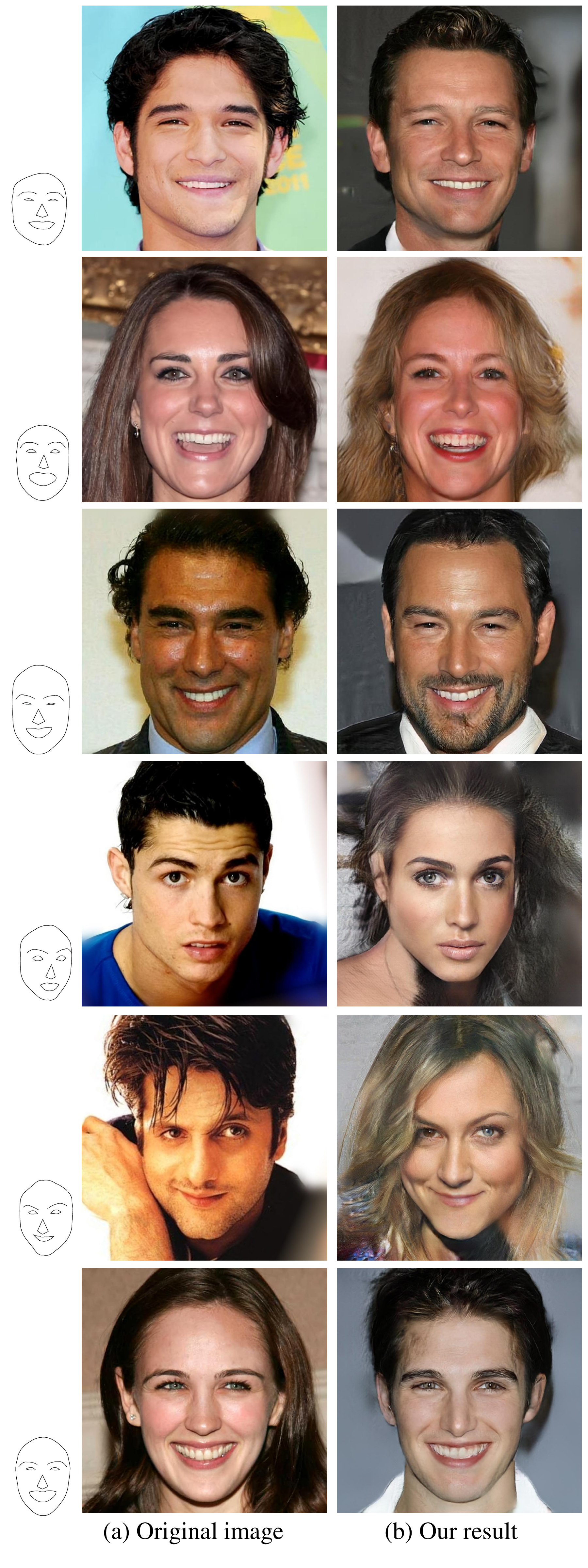}  
  \caption{Example edge-to-face results on the CelebA-HQ dataset~\cite{karras2017progressive} (edge maps shown at lower left corners).
  }
  \lblfig{edge2face}
\end{figure}

The results in this paper suggest that conditional GANs can synthesize high-resolution photo-realistic imagery without any hand-crafted losses or pre-trained networks. We have observed that incorporating a perceptual loss~\cite{johnson2016perceptual} can slightly improve the results. Our method allows many applications and will be potentially useful for domains where high-resolution results are in demand but pre-trained networks are not available (e.g.,\ medical imaging~\cite{guibas2017synthetic} and biology~\cite{costa2017end}).

This paper also shows that an image-to-image synthesis pipeline can be extended to produce diverse outputs, and enable interactive image manipulation given appropriate training input-output pairs (e.g.,\ instance maps in our case). Without ever been told what a ``texture'' is, our model learns to stylize different objects, which may be generalized to other datasets as well (i.e.,\ using textures in one dataset to synthesize images in another dataset).
We believe these extensions  can be potentially applied to other image synthesis problems.

\vspace{.15in}
{\noindent \bf Acknowledgements} We thank Taesung Park, Phillip Isola, Tinghui Zhou, Richard Zhang, Rafael Valle and Alexei A. Efros for helpful comments. We also thank \ck and Isola et al.~\cite{isola2016image} for sharing their code. JYZ is supported by a Facebook graduate fellowship.

\clearpage{\thispagestyle{empty}\cleardoublepage}
\bibliographystyle{ieee}
\bibliography{main}
\clearpage 

\appendix
\section{Training Details}
All the networks were trained from scratch, using the Adam solver~\cite{kingma2014adam} and a learning rate of $0.0002$. We keep the same learning rate for the first $100$ epochs and linearly decay the rate to zero over the next $100$ epochs. Weights were initialized from a Gaussian distribution with mean $0$ and standard deviation $0.02$. We train all our models on an NVIDIA Quadro M6000 GPU with $24$GB GPU memory.

The inference time is between $20\sim 30$ milliseconds per $2048\times1024$ input image on an NVIDIA 1080Ti GPU with $11$GB GPU memory. This real-time performance allows us to develop interactive image editing applications. 

Below we discuss the details of the datasets we used. 

\begin{itemize}
    \item {\bf Cityscapes dataset~\cite{Cordts2016cityscapes}: } $2975$ training images from the Cityscapes training set with image size $2048 \times 1024$. We use the Cityscapes validation set for testing, which consists of 500 images.
    
    \item {\bf NYU Indoor RGBD dataset~\cite{Silberman2012indoor}: }
$1200$ training images and $249$ test images, all at resolution of $561\times 427$.

    \item {\bf ADE20K dataset~\cite{zhou2017scene}: }
$20210$ training images and $2000$ test images with varying image sizes. We scale the width of all images to $512$ before training and inference.

    \item {\bf Helen Face dataset~\cite{le2012interactive,smith2013exemplar}: }
$2000$ training images and $330$ test images with varying image sizes. We resize all images to $1024\times 1024$ before training and inference.
\end{itemize}

\section{Generator Architectures}
Our generator consists of a global generator network and a local enhancer network. we follow the naming convention used in Johnson el al.~\cite{johnson2016perceptual} and CycleGAN~\cite{zhu2017unpaired}. Let \texttt{c7s1-k} denote a $7\times7$ Convolution-InstanceNorm~\cite{ulyanov2016instance}-ReLU layer with $k$ filters and stride $1$. \texttt{dk} denotes a $3\times3$ Convolution-InstanceNorm-ReLU layer with $k$ filters, and stride $2$. We use reflection padding  to reduce boundary artifacts. \texttt{Rk} denotes a residual block that contains two $3\times3$ convolutional layers with the same number of filters on both layers. \texttt{uk} denotes a $3\times3$ fractional-strided-Convolution-InstanceNorm-ReLU layer with $k$ filters, and stride $\frac{1}{2}$.

Recall that we have two generators: the global generator and the local enhancer.  

Our global network:\\
\texttt{c7s1-64,d128,d256,d512,d1024,R1024,R1024,\\
R1024,R1024,R1024,R1024,R1024,R1024,R1024,\\
u512,u256,u128,u64,c7s1-3}

Our local enhancer:\\
\texttt{c7s1-32,d64\footnote{We add the last feature map (\texttt{u64}) in our global network to the output of this layer.},R64,R64,R64,u32,c7s1-3}

\section{Discriminator Architectures}
For discriminator networks, we use $70\times 70$ PatchGAN~\cite{isola2016image}. 
Let \texttt{Ck} denote a $4\times4$ Convolution-InstanceNorm-LeakyReLU layer with k filters and stride $2$. After the last layer, we apply a convolution to produce a $1$ dimensional output. We do not use InstanceNorm for the first \texttt{C64} layer. We use leaky ReLUs with slope $0.2$. All our three discriminators have the identical architecture as follows:\\

\texttt{C64-C128-C256-C512}

\section{Change log}
\paragraph{v1} initial preprint release
\paragraph{v2} CVPR camera ready, adding more results for edge-to-photo examples.

\end{document}